\documentclass{article}
\usepackage{colm2024_conference}

\pdfoutput=1

\usepackage{microtype}
\usepackage[hyperfootnotes=false]{hyperref}
\usepackage{url}
\usepackage{booktabs}
\usepackage{graphicx}
\usepackage{listings}
\usepackage{enumitem}
\usepackage{color}
\usepackage{xspace}
\usepackage{bm}
\usepackage{xcolor}

\usepackage{multirow}
\usepackage{makecell}
\usepackage{siunitx}
\usepackage{etoolbox}
\robustify\bfseries

\usepackage{tabularx}

\usepackage[bottom]{footmisc} 

\newcommand{\expansionsize}{{\emph{expansion~size}}\xspace}
\newcommand{\groupsize}{{\emph{group~size}}\xspace}
\newcommand{\budget}{{\emph{labeling~budget}}\xspace}

\title{Can LLMs get help from other LLMs\\ without revealing private information?}

\author{Florian Hartmann\thanks{Corresponding author.} \\
Google Research\\
\texttt{fhartmann@google.com} 
\And
Duc-Hieu Tran \\
Google Research\\
\texttt{hieuza@google.com} 
\And
Peter Kairouz \\
Google Research\\
\texttt{kairouz@google.com}
\AND
Victor Cărbune \\
Google Research\\
\texttt{vcarbune@google.com} 
\And
Blaise Aguera y Arcas \\
Google Research\\
\texttt{blaisea@google.com} 
}

\colmfinalcopy

\begin{document}

\maketitle

\begin{abstract}
Cascades are a common type of machine learning systems in which a large, remote model can be queried if a local model is not able to accurately label a user’s data by itself. Serving stacks for large language models (LLMs) increasingly use cascades due to their ability to preserve task performance while dramatically reducing inference costs. However, applying cascade systems in situations where the local model has access to sensitive data constitutes a significant privacy risk for users since such data could be forwarded to the remote model. In this work, we show the feasibility of applying cascade systems in such setups by equipping the local model with privacy-preserving techniques that reduce the risk of leaking private information when querying the remote model. To quantify information leakage in such setups, we introduce two privacy measures. We then propose a system that leverages the recently introduced social learning paradigm in which LLMs collaboratively learn from each other by exchanging natural language. Using this paradigm, we demonstrate on several datasets that our methods minimize the privacy loss while at the same time improving task performance compared to a non-cascade baseline.
\end{abstract}

\section{Introduction}

% Putting this here to have a visualization on the first page.
\begin{figure*}[b]
    \centering
    \includegraphics[width=\linewidth]{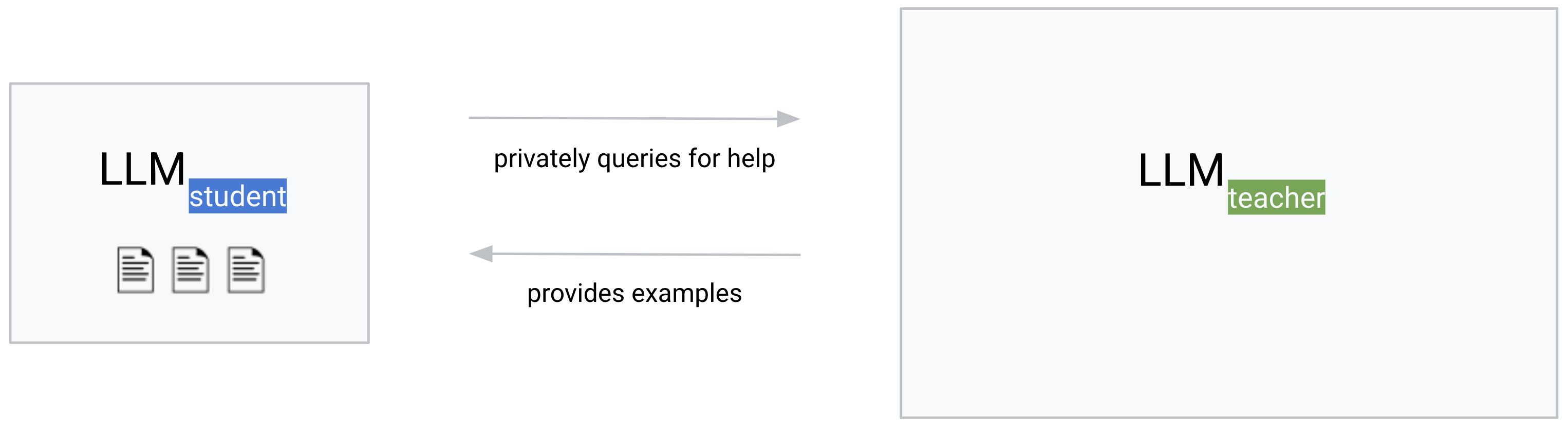}
    \caption{The local model, the \textit{student}, wants to label its private data. It can query a larger, remote model, the \textit{teacher}, to get help. The student may not reveal private data to the teacher.}
    \label{fig:system}
\end{figure*}

% Motivation: the best LLMs run far from the local context that private data exists in.
Large language models (LLMs) such as Gemini Ultra by \citet{team2023gemini} and GPT-4 by \citet{openai2023gpt4} are reporting remarkable performance on many tasks. These models, however, not only come with high inference costs, but they also have to run in data centers far from the local contexts where private data is available. Conversely, models that can run in private contexts, such as Gemini Nano, have more limited capabilities since they run on the user’s device.

% Cascades enable closely matching performance but do not work with private data.
To unlock state-of-the-art performance in private contexts, local models with access to sensitive data need to be equipped with a privacy-preserving mechanism that enables querying a remote model without sharing any sensitive data. Although standard cascade systems in which a smaller, less capable model, queries a larger, much more capable one in order to solve a task have previously been studied \citep{yue2024large, chen2023frugalgpt}, privacy-preserving ones have not yet been explored. In today's cascade systems, the decision of whether a larger model should be leveraged or not is usually done through an additional mechanism that determines whether the query can be handled by the smaller model independently  \citep{li2021cascadebert}. If determined to be handled by the larger model, the query is simply forwarded without consideration for the private data it may contain. This poses privacy threats for users, ranging from leaking sensitive data to the forwarded sample even being ingested in training datasets of the remote system.

% Our goal is a privacy-preserving cascade system.
We introduce the first privacy-preserving approach to cascade systems.
Contrasting to standard cascade systems, our local model always assumes its data is private.
As such, the local model should not share anything private with the remote model.
Going one step further, even if the local model does not verbatim share private information, we aim to prevent a curious remote model operator from reconstructing private data by utilizing auxiliary information it might have.
To focus on these challenges, we assume there are no efficiency constraints and that the local model can always ask for help from the remote model, as shown in Figure~\ref{fig:system}.
Therefore our optimal cascade setup consists of minimizing the privacy loss while maximizing task performance, where an upper bound is given by querying the teacher model with the actual data, although private.

% To implement such a system, we use ICL and social learning.
To succeed at this task, the local model, typically smaller and less capable, needs to find the right balance between revealing sufficient information about the problem to receive useful signals from the more capable, remote model while keeping details private.
To enable learning from the remote model, the local model makes use of gradient-free learning capabilities through natural language that in-context learning (ICL) capabilities of LLMs enable~\citep{brown2020language}.
Throughout, we leverage the recently introduced social learning paradigm by~\cite{mohtashami2023social, bandura1977social} in which LLMs learn through natural language from other LLMs.

\paragraph{Contributions} We summarize our contributions as follows: \textbf{(i)}~we enable cascade systems to be used where access to private data is necessary to solve a task, but cannot be revealed \textbf{(ii)}~we design and evaluate algorithms that sanitize private data while still leveraging in-context learning capabilities of private models and \textbf{(iii)}~whereas previous work to the best of our knowledge analyzes settings without auxiliary information, we go one step further by considering auxiliary information and proposing a novel metric to this end  \textbf{(iv)}~we perform extensive experiments on a diverse range of tasks, quantifying task performance and impact on privacy using standardized measures.

\section{Problem Setting}

% Teacher and student
Our paper considers a variant of social learning~\citep{mohtashami2023social} where neither of the participants has any labeled data. A local model, called the \emph{student}, has private data that it cannot label well by itself. A larger, remote model, called the \emph{teacher}, can do a better job at labeling the data. These two models form a cascade, in which the student can improve its performance by communicating with the teacher. We call what the student sends to the teacher a \emph{query}. Figure~\ref{fig:system} shows a visualization of this setup.

% Constraints on communication
\paragraph{Constraints} There are two constraints on the queries from student to teacher. \textbf{(i)} The communication must be privacy-preserving, i.e. the student may not copy over its data and must not reveal anything private. \textbf{(ii)} There is only a single round of communication between student and teacher, meaning neither of them can maintain any state or update a model of the other's capabilities.

% 3 step iterative. (1) query, (2) respond, (3) learn in-context
To this end, all algorithms follow the same structure. Given (0) that the student needs help, it then (1) uses it's private data to generate a query to the teacher. In turn, (2) the teacher uses the query to generate ICL examples for the student. Finally, (3) the student uses the ICL examples to go back to solving its original problem.

% Simplifying assumption 1: always ask for help
\paragraph{Simplifying assumptions} To better focus on the challenges we aim to address in this paper, we furthermore make two simplifying assumptions. \textbf{(i)} We assume that communication with the remote teacher is always helpful. This assumption is reasonable because existing techniques for determining delegation in cascades, discussed in Section~\ref{sec:background}, could be combined with our methods.
% Simplifying assumption 2: template
\textbf{(ii)} We also assume that both student and teacher are aware of the format of the examples, as shown in Table~\ref{fig:template} in the appendix. Such an assumption is useful because we want the student to learn more complex things about the data from the teacher instead of simply learning a format or chain of thought prompt.

% Goal
Given this problem setting, the goal of the student is to maximize its performance in correctly labeling its data while not revealing anything private that is part of said data.

\section{Privacy Measures}
\label{sec:privacy}

The student's data may often contain sensitive personal information that should be kept hidden from an untrusted, or partially trusted, teacher. For example, consider a query that tries to figure out what disease could best explain a set of health symptoms that are experienced by a user after they have engaged in a specific sequence of activities. Here, being able to associate the set of activities and/or symptoms with a specific user is a privacy violation that we would like to eliminate. 

To address such privacy  violations, one might be tempted to resort to data anonymization techniques, such as \emph{differential privacy} (DP)~\citep{dwork2006differential}. However, these techniques are most useful when computing aggregates across many users (e.g. average of gradient vectors computed on a batch of sensitive training examples). Using the local model of DP~\citep{warner1965randomized, evfimievski2003limiting, kasiviswanathan2011can} as a mechanism to mask private information in the query will end up masking both private and non-private information in the query, rendering the masked query useless for the task at hand. Alternatively, using the ICL model of DP~\citep{liu2024prompt, wu2023privacy} to privatize the sensitive parts of the student's data suffers a major hurdle: it assumes the student has many private examples it can jointly consider when creating a query to the teacher. While we do look into grouping examples when generating a query in Section~\ref{sec:grouping}, we expect the student to have very few private examples, and want it to be able to generate privacy-preserving queries even when only having a single, private example. DP-ICL cannot work in such a setting.

Instead, we leverage data minimization privacy techniques, specifically \emph{contextual integrity}~\citep{nissenbaum2004privacy} which describes privacy as an appropriate flow of information. Under this technique, the student would keep information that is useful for the task (e.g. the activities \& symptoms in the above-mentioned example) but remove any personally identifying information that is not relevant to the query context. We note that even under perfect masking, this approach could still leak sensitive information, should the teacher model have access to auxiliary information that can be used to identify certain unique features that are strongly correlated with the “perfectly masked” prompts~\citep{narayanan2008robust,sweeney2002k}. Thus, an important contribution of our work is a methodology for measuring and assessing leakage under auxiliary information. 

The success of our approach hinges on correctly identifying and masking the sensitive parts of the query without tampering with the description of the task. To this end, we propose various techniques that can analyze information in queries to produce safe queries that can be shared with the teacher model. To assess the privacy of the queries, we consider two concrete metrics, the \emph{entity leak} metric that counts entities that exist in both original examples and the student's queries, and the \emph{mapping leak} metric that considers a setting with auxiliary information.

\paragraph{Entity leak metric} Contextual integrity states that privacy is the appropriate flow of information. For most production applications, it is hard to say what is appropriate to share. As a proxy for this, we consider the interpretable metric of leaked entities. All entities, such as names, locations, email addresses, or numbers, in the dataset, are considered to be private. We measure how many of the entities in the original example are still part of the student's query upon masking.

\paragraph{Mapping leak metric} Even if all entities are removed from the student's query, it is still possible for a curious teacher to reconstruct private information by carefully analyzing the query. Indeed, auxiliary information that the teacher may have access to can help it be more effective at this. We measure how well the teacher could do this through a worst-case analysis. More precisely, we assume the teacher is presented with 1 original example and 100 masked queries out of which exactly one was generated from the original example. We measure how often the teacher is able to correctly map the original example to this particular (masked) query out of the 100 options. Providing the teacher with a complete original example represents an upper bound on the auxiliary information the teacher could have. To do a better job at this mapping, we allow the teacher to query the student model, which is useful since it was used to generate the masked query. To conduct the mapping, we then score continuations of the original example and the 100 generated queries, and measure how often the correct query scores the highest. We show that access to such (worst-case) auxiliary information could lead to non-trivial privacy leakage even when the entities are properly masked. 

\section{Methods}
\label{sec:methods}

\begin{figure*}[h]
    \centering
    \includegraphics[width=\linewidth]{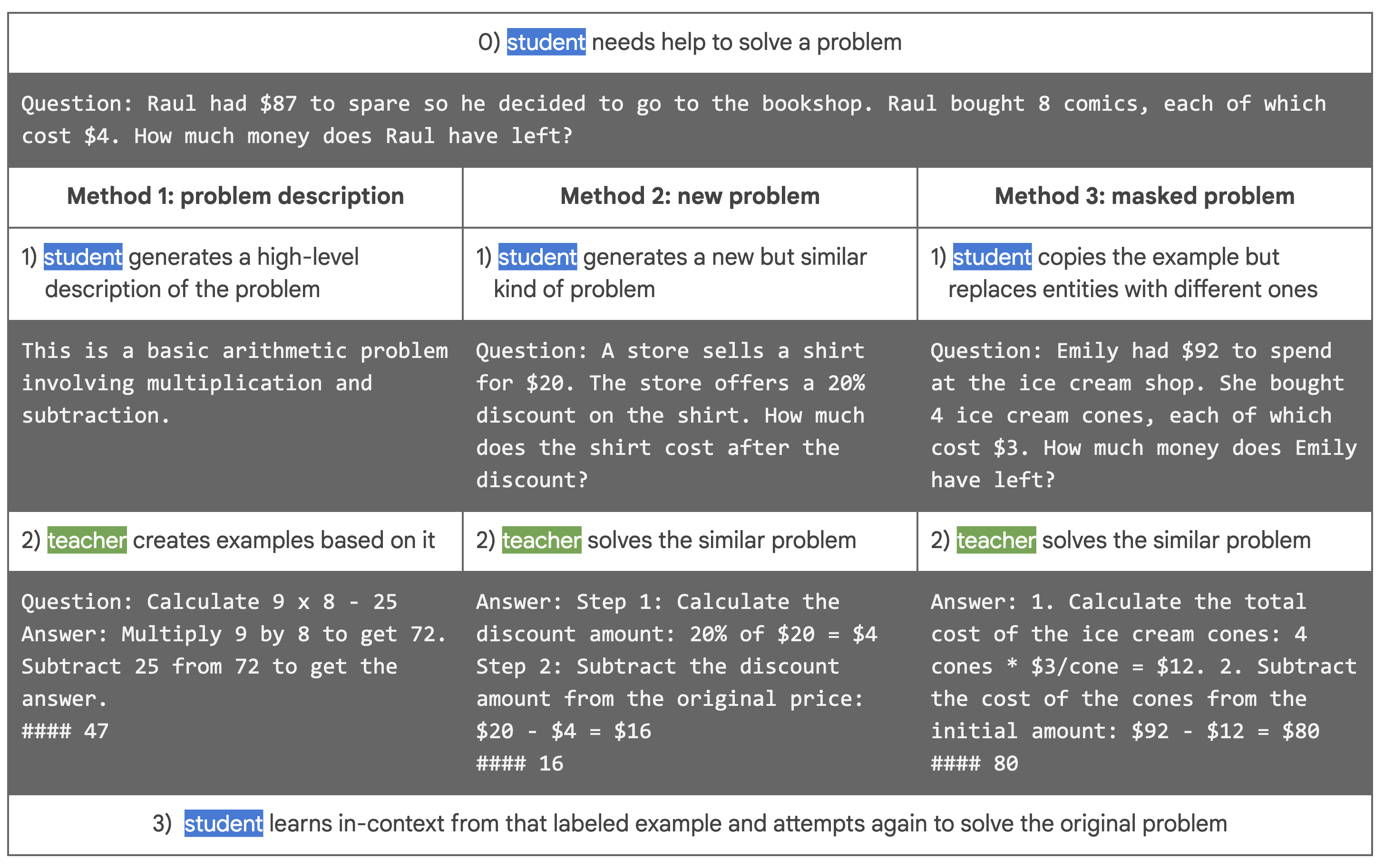}
    \caption{The three methods we consider. Steps 1 and 2 show actual student queries and teacher responses as generated in our experiments when using Gemini 1.0 Nano-2 as the student and Gemini 1.0 Ultra as the teacher. Note that each method generates increasingly specific queries about the student's problem.}
    \label{fig:algs}
\end{figure*}

% Three methods based on the problem statement
We consider three algorithms for how the student could privately learn from the teacher, as shown in Figure~\ref{fig:algs}.
The first of these methods is based on the student describing the problem it is facing while the latter two methods generate similar, non-private examples that the teacher can label.
As a hyperparameter for all these methods, we consider the \expansionsize to denote how many labeled ICL examples the student will receive from the teacher.

\subsection{Method 1: Creating a problem description}

% Student: Generates high-level description
As an initial approach, we consider a method in which the student analyzes the problem it is given and generates a high-level description from this problem.
Even if the student cannot solve the problem, it might be able to describe the type of problem it is facing.
This description is the query to the teacher.

% Teacher: Creates few-shot examples
The teacher in turn wants to create few-shot examples that the student can use to solve the problem it is facing.
Since the teacher has access to a template about the example structure, it knows what format to follow.
To create such examples, it then uses this template as well as the student's description to create \expansionsize many new examples.

\subsection{Method 2: Generating new unlabeled examples}
\label{sec:generation}

% New examples can be private but just as educational
Instead of providing the teacher with an abstract description of the problem it is facing, the student can generate a similar, but novel problem itself. As a motivating example for why this is a sensible choice, consider GSM8k~\citep{cobbe2021training}, a math dataset with problems of US middle school difficulty. Given such a math problem, it is possible to create a similar math problem that is just as educational but contains none of the same details, i.e. both problems follow a similar structure and are of similar difficulty.

% LLMs can generate new examples
Previous work has shown that LLMs are able to generate new examples from original examples that they see in-context~\citep{shao2023synthetic,mohtashami2023social}. We additionally observe that for many tasks it is easier to generate new examples than it is to solve them, meaning it is possible for the student model to synthesize similar, unlabeled examples, even if it does not do a good job at labeling them.

% Structure of method 2
To this end, our second method works as follows: We (1) prompt the student LLM to generate  \expansionsize new unlabeled examples. Then, (2) the teacher receives these examples and labels them. Finally, (3) the student learns in-context from that and tries to solve the original problem. Throughout, both teacher and student models  utilize the task template to understand what format the labeled and unlabeled examples follow and for where step-by-step explanations make sense.

\subsection{Method 3: Replacing entities in original examples}
\label{sec:masking}

% Mask instead of generate new examples
Instead of instructing the student to generate completely novel examples, we can also ask it to keep the same example while replacing names, locations, numbers, and other entities. The student then generates a new unlabeled example that is very similar to the original but that contains none of the private information.
Since there are many ways to replace the entities, we can again generate \expansionsize examples using this technique.

% By prompting
While this could be done using a specialized entity detection model and rule-based systems, we observe that LLMs do a fairly good job at this themselves. Thus, we decide to simply prompt the student model to find and replace private entities.
% Structure of method 3
The full flow of this method is the same as the method in Section~\ref{sec:generation}, except that in step (1), we now simply replace entities instead of generating completely new examples.

\subsection{Grouping unlabeled examples to reduce teacher calls}
\label{sec:grouping}

% Teacher is an expensive resource
Each call to the teacher implies some chance of leaking private information. This chance needs to be traded off with how much the student can be improved through this process. Like in active learning, the teacher in our setting can thus be considered an expensive resource that needs to be used economically.

% Introducing group size
To utilize this resource well, we introduce an additional hyperparameter \groupsize that denotes how many private examples the student groups together in order to create \expansionsize many ICL examples through the teacher. The student considers the entire group jointly when synthesizing descriptions and new unlabeled examples, and is thus able to combine information from the grouped, private examples. By \budget = \expansionsize / \groupsize, we denote the budget of how many teacher labeled examples may be created for each original example.
Note that the student does not get to choose which examples to group together.

\section{Experiments}

In order to evaluate the effectiveness of the methods introduced in Section~\ref{sec:methods}, we evaluate them in terms of accuracy and privacy on a diverse group of datasets and compare them against two baselines.

% Gemini models, normalized by teacher performance
\paragraph{Models} We use the \textit{Gemini 1.0} family of models~\citep{team2023gemini} for all of our experiments. As the teacher, we utilize \textit{Ultra}, the most powerful model of the family. In most of our experiments, \textit{Nano-2}, a 3.5B parameter model that can be deployed on mobile phones, is the student. The student model capabilities naturally influence the performance of our method and hence we  also run experiments when \textit{Pro} is the student. In line with previous reports on Nano's performance~\citep{team2023gemini}, we normalize task success in all our experiments by the teacher's performance since it is an upper bound for what we can hope to achieve.

% Datasets: GSM8k, MT, intent recognition
\paragraph{Datasets} We consider a variety of datasets in our experiments to demonstrate that our methods generalize across a suite of tasks: GSM8k math problems~\citep{cobbe2021training}, assistant intent detection~\citep{srivastava2022beyond}, classifying whether statements are subjective or objective~\citep{conneau2018senteval} and mid-resource machine translation~\citep{tiedemann2020tatoeba}. See Appendix~\ref{app:datasets} for a more detailed description of the datasets.

% Private entities as detected by Ultra
%\paragraph{Private entities} Contextual integrity, the notion of privacy we follow, states that privacy is the appropriate flow of information. What is appropriate depends on the application and is hard to say in the absence of a product application. To overcome this, we declare all entities (such as names, locations, and numbers) that appear in datasets to be private.

\paragraph{Baselines} We compare against a \textit{weak} and \textit{strong} baseline. For the \emph{weak baseline}, we consider a student that does not communicate with the teacher at all. Since the student does not have any labeled data on its own, it thus falls back to the 0-shot setting while still being able to use the task's template. As the \emph{strong baseline}, we evaluate a student that has access to 8 arbitrary, golden examples. We consider this to be a strong baseline since these examples are perfectly labeled and for the same task that the student is trying to solve. In practice, such data does often not exist and cannot be easily matched to the student's problem.

\subsection{Task Performance}

% Here's table 1. We do well, but the student matters
To evaluate our methods, we run experiments for all above mentioned datasets. For ease of comparison, we consider the 8-shot performance of each method. Table~\ref{tab:main-utility} shows these results. Across all datasets, we outperform both the weak and strong baseline. However, we note that for GSM8k, getting close to 100\% task success, as normalized by the teacher's performance, requires Pro as a strong student model.

% Result 1: big table with the best results per dataset
\newcommand{\smallbold}[1]{\small{\textbf{#1}}}
\begin{table*}[h]
    \centering
\scalebox{0.85}{
    \begin{tabular}{c|c||c|c||c|c|c}\toprule
    \smallbold{Dataset} &
    \smallbold{Student} &
    \makecell{\smallbold{Weak Baseline:}\\\smallbold{0-shot}} &
    \makecell{\smallbold{Strong Baseline:}\\\smallbold{Golden Data}\\\smallbold{8-shot}} &
    \makecell{\smallbold{Method 1:}\\\smallbold{Descriptions}\\\smallbold{8-shot}} &
    \makecell{\smallbold{Method 2:}\\\smallbold{New Problems}\\\smallbold{8-shot}} &
    \makecell{\smallbold{Method 3}:\\\smallbold{Replacing}\\\smallbold{8-shot}} \\\midrule
    
\multirow{2}{*}{\small{GSM8k}} & \small{Nano-2} & 11.3\% & 34.9\%& 36.7\% & 45.6\% & \textbf{55.9\%}  \\
& \small{Pro} & 85.4\% & 91.1\% & 78.6\% & 91.6\% & \textbf{98.3\%} \\
\midrule

\makecell{\small{Intent}\\\small{Recognition}} & \small{Nano-2} & 70.9\% & 92.4\% & 82.7\% & 92.3\% & \textbf{94.6\%} \\
%& Pro & 97.8\% & \textbf{98.6\%} & 97.0\% & 97.3\% & 98.3\%  \\
\midrule

\Gape[1ex][1ex]{\small{Subj}} & \small{Nano-2} & 55.6\% & 74.2\% & 74.2\% & 71.0\% & \textbf{79.7\%} 
\\
\midrule

\makecell{\small{Translation}\\\small{en $\rightarrow$ eu}} & \small{Nano-2} & 70.8\% & 72.9\%  &	72.8\% & 74.8\% & \textbf{91.0\%} \\
\bottomrule
    \end{tabular}
}
    \caption{Task performance with Gemini 1.0 Nano-2 and Pro as students, and Gemini 1.0 Ultra as the teacher. All values are normalized by the teacher's performance as reported in Table~\ref{tab:teacher-utility}.  For easier comparison, we only consider setups with \expansionsize~$ = 8$, \groupsize~$ = 1$ here. Note that we report BLEURT~\citep{sellam2020bleurt} for machine translation and accuracy for all other tasks. Appendix~\ref{app:mt-performance} shows similar machine translation results for 6 more languages.}
    \label{tab:main-utility}
\end{table*}

% Why method 3 performs so well and method 1 so poorly
We observe that method 3 performs very well across all datasets. Likely this is because the queries generated by this method are the closest to the problem that the student aims to solve. Method 1 performs the worst. We find this method to be the hardest to get to work well since for some tasks, e.g.\ intent recognition, the student model is only able to explicitly describe the unlabeled example if it is also able to label it, rendering it a less competitive method.

\begin{figure*}[h]
    \centering
    \includegraphics[width=.5\linewidth]{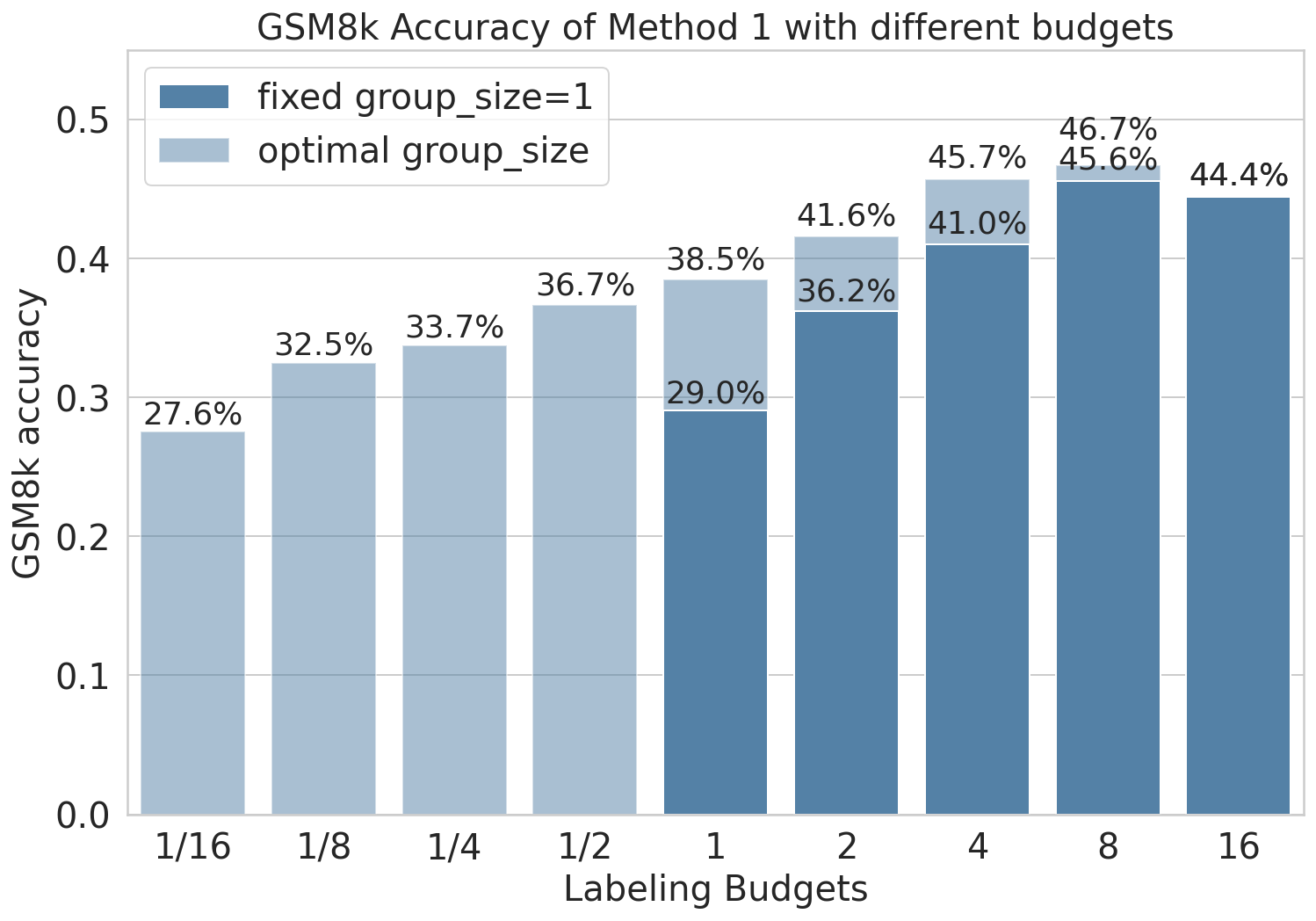}
    \caption{For a given \budget = \expansionsize / \groupsize, we show the accuracy reached. Grouping allows us to improve the $29.0\%$ accuracy reached through \expansionsize $=$ \groupsize~$= 1$ to an accuracy of $32.5\%$ all while just using $\frac18$ of the budget. Furthermore, even for budgets above 1, we can well outperform the approach without grouping.}
    \label{fig:budget}
\end{figure*}

% Budgeting results.
Furthermore, to investigate the best use of the \budget = \expansionsize / \groupsize, we run full grid searches for the different methods. For each \budget, we then obtain the best performance that can be reached. As shown in Figure~\ref{fig:budget}, the choice of these hyperparameters allows budgets below 1, which is not possible without grouping.

% Result 2: bar plot for where each bar is a different budget

\subsection{Privacy Results}

% No auxiliary information: leakage is close to 0, but occasionally things do indeed leak
To analyze how our methods fare in terms of privacy, we compute the two metrics mentioned in Section~\ref{sec:privacy}. We find entities in both the original example and in the query by asking Gemini 1.0 Ultra to play an entity detector that finds entities such as names, locations, numbers, and anything else one might consider private. We manually verify on a subset of examples that this does indeed find the desired entities. The results in Table~\ref{tab:main-privacy} show the results of this analysis.

\newcommand{\smallerbold}[1]{\footnotesize{\textbf{#1}}}
% Result 3: table with auxiliary information leakage
\begin{table*}[h]
    \centering
\scalebox{.8}{
    \begin{tabular}{c|c||c|c|c|c|c}\toprule
    \smallbold{Dataset} &
    \smallbold{Metric} &
   \makecell{\smallbold{Method 1:}\\\smallbold{Descriptions}\\\smallbold{8-shot}\\ \smallerbold \groupsize $\bm{ = 1}$} &
    \makecell{\smallbold{Method 2:}\\\smallbold{New Problems}\\\smallbold{8-shot}\\ \smallerbold \groupsize $\bm{ = 1}$} &
    \makecell{\smallbold{Method 2:}\\\smallbold{New Problems}\\\smallbold{8-shot}\\ \smallerbold \groupsize $\bm{ = 2}$} &
    \makecell{\smallbold{Method 2:}\\\smallbold{New Problems}\\\smallbold{8-shot}\\ \smallerbold \groupsize $\bm{ = 4}$} &
    \makecell{\smallbold{Method 3}:\\\smallbold{Replacing}\\\smallbold{8-shot}\\ \smallerbold \groupsize $\bm{ = 1}$}
\\\midrule
    
\multirow{3}{*}{\small{GSM8k}} & \small accuracy & 36.7\% & 45.6\% & 45.7\% & 40.2\% & \textbf{55.9\%} \\
& \small entity leaks & 2.7\% & 1.5\%  & 0.7\% & \textbf{0.2\%} & 1.2\%   \\
& \small mapping leak & 16.4\% & 5.4\% & 2.9\% & \textbf{2.5\%} & 53.8\%  \\
\midrule

\multirow{3}{*}{\makecell{\small{Intent}\\ \small{Recognition}}} & \small accuracy & 82.7\% & 92.3\% & 89.9\% & 86.7\% & \textbf{94.6\%} \\
& \small entity leaks & 5.7\% & 3.9\% & 0.7\% & \textbf{0.1\%} & 0.5\% \\ 
& \small mapping leak & 1.8\% & 2.6\% & 1.0\% & \textbf{0.9\%} & 3.2\% \\
\midrule

\multirow{3}{*}{\makecell{\small{Subj}}} & \small accuracy & 74.2\% & 71.0\% & 64.2\% & 61.4\% & \textbf{79.7\%} \\
& \small entity leaks   & 4.3\% & 3.8\% & 1.3\% & \textbf{0.6\%} & 1.4\% \\ 
& \small mapping leak  & 16.6\% & 6.2\% & 3.2\% & \textbf{2.8\%} & 43.3\%\\
\midrule

\multirow{3}{*}{\makecell{\small{Translation}\\\small{en $\rightarrow$ eu}}} & \small BLEURT & 72.8\% & 74.8\% &	68.2\% & 69.0\% & \textbf{91.0\%} \\
& \small entity leaks  & 2.5\% & 1.3\% & 1.3\% & \textbf{0.0\%} & 1.3\% \\ 
& \small mapping leak  & 4.5\% & 3.5\% & 1.3\% & \textbf{1.2\%} & 2.9\% \\
\bottomrule
    \end{tabular}
}
    \caption{For Nano-2 as the student, and each dataset and method, we present our two privacy metrics. Method 3 generally achieves the best quality results while leaking few entities. Method 2 with grouping offers the strongest privacy metrics.}
    \label{tab:main-privacy}
\end{table*}

% Entity leakage amount: method 1 > 2 > 3
Unexpectedly, we observe that method 1 often leaks the most entities. While this method should generate the most high-level queries in theory, it is hard to get to work well in practice. On a subset of original examples, the student is not able to synthesize a high-level description and instead defaults to detailedly describing the problem it is facing. While the queries of method 3 are the closest to the original messages, they also leak the fewest entities. We hypothesize that this is because the student does not need to understand the problem well in order to find and replace entities. It can simply consider the individual tokens and replace them without needing to understand what kind of problem it is trying to get help on.

% Auxiliary information: Masking does badly. 
However, when analyzing the mapping metric, which describes a worst case of how well an attacker with auxiliary information can identify original examples, the results paint a different picture. Here, method 3 performs significantly worse. While few entities leak in this method, the structure and writing style are maintained, making it especially easy to map between original and generated example. This is in particular the case for the GSM8k and Subj datasets in which examples have a distinct structure that makes them easy to identify.

% grouping helps.
We find grouping examples to work particularly well with method 2. We observe that with \groupsize $ = 2$ leaks in both metrics significantly reduce, and in the case of GSM8k even without a drop in performance.

% Instruction and grouping is useful.
Finally, we note that the choice of the right method depends on the concrete threat model considered. While method 1 is neither convincing in terms of quality and privacy, method 3 works remarkably well in situations where the threat model does not involve auxiliary information. Conversely, if one does consider auxiliary information, method 2, potentially with grouping, is the most appropriate to use.

\subsection{Qualitative Analysis}
\label{sec:qualitative}

% We analyze errors automatically
In order to better understand where our methods help and where they fall short, we run detailed analyses on the predictions that the student model is able to make after it got help from the teacher. To do this at scale, we ask Gemini 1.0 Ultra to look at the golden label and the student's prediction, and classify the errors into certain classes. We confirm manually for a subset of cases that these classifications make sense.
% GSM8k table is here but the appendix has more
Table~\ref{table:qualitative} shows the results of this analysis for GSM8k based on 500 examples, for the strong baseline and the best setup for each of our methods.
We show similar analyses for machine translation in Appendix~\ref{app:mt-qualitative}, as well as example student queries for all datasets in Appendix~\ref{app:examples}.

% Result 4: table with GSM8k error classes
\begin{table*}[h]
\centering

\def\arraystretch{1.5}
\scalebox{.85}{
\begin{tabular}[h]{c||c||c|c|c} 
\toprule
 \smallbold{Class} &
 \makecell{\smallbold{Strong Baseline:}\\\smallbold{Golden Data}\\\smallbold{8-shot}} &
    \makecell{\smallbold{Method 1:}\\\smallbold{Descriptions}\\\smallbold{8-shot}} &
    \makecell{\smallbold{Method 2:}\\\smallbold{New Problems}\\\smallbold{8-shot}} &
    \makecell{\smallbold{Method 3}:\\\smallbold{Replacing}\\\smallbold{8-shot}}
 \\
\hline \hline
\small Correct prediction & 29.0\% & 32.2 \% & 40.0\% & \textbf{49.1\%} \\  \hline  \hline
\small Calculation error & 42.8\% & 35.7\% & 32.2\% & \textbf{26.8\%} \\ \hline
\small Flaw in reasoning & 13.8\% & 10.8\% & 11.4\% & \textbf{8.1\%} \\ \hline
\small Using incorrect information & \textbf{5.3\%}& 9.1\% & 6.4\% & 6.5\%  \\ \hline
\small Incorrectly applying formulas & 4.9\% & 5.1\% & 5.6\% & \textbf{4.3\%} \\ \hline
\small Not understanding the problem & \textbf{4.3}\% & 7.1\% & 4.3\% & 5.2\% \\
\bottomrule
\end{tabular}
}
\caption{An analysis of the student's predictions shows that calculation and reasoning errors of the students reduce through the ICL examples our method provides. Errors caused by using incorrect information slightly increase, likely because the student model can get confused by the similar examples it is seeing. We bold the best cell of each row to emphasize that method 3 shows the most impressive reduction in mistakes. Note that we do not normalize by teacher task success here as opposed to the other tables.}
\label{table:qualitative}
\end{table*}

\section{Related Work}
\label{sec:background}

\paragraph{LLM Cascades} Cascades were mostly studied for improving overall inference costs, particularly given ever-increasing LLM sizes~\citep{hoffmann2022training}. Task performance steadily increases with parameter count~\citep{schaeffer2023emergent}. Various approaches to cascade inference are compared in \cite{miao2023efficient}. Some methods~\citep{li2021cascadebert, chen2023frugalgpt} use a classifier to determine whether to forward a query or not, while more recent work~\citep{yue2024large} leverages a voting and consistency measure of the first model in the cascade as proxy for the inability to provide an answer. % \cite{dohan2022language} call out a general probabilistic framing to cascades.
We replaced inference cost with a privacy measure optimization and quantified to what degree task performance can be preserved.
\paragraph{Differential Privacy (DP)} DP formalizes privacy guarantees in a probabilistic framework~\citep{dwork2006differential}. This can be implemented in various ways, e.g.\ via the local model of DP~\citep{warner1965randomized,evfimievski2003limiting,kasiviswanathan2011can} or as part of in-context learning~\citep{liu2024prompt,wu2023privacy}. While these techniques are useful when computing aggregates across many users, we want our system to work even when a user only has a single, private example, as explain in Section~\ref{sec:privacy}.
\paragraph{Data Minimization} As an alternative to DP, we follow data minimization principles in the form of contextual integrity~\citep{nissenbaum2004privacy}. Data minimization techniques are particularly important for removing sensitive information from LLM training datasets. \cite{lison-etal-2021-anonymisation} present an overview of many techniques relevant to enabling cascade systems in private/public setups. In this work, we investigated the effectiveness of masking operations, and instead of using a separate sequence tagging model we relied on the student LLM capability to perform such transformations. Recent studies, such as~\cite{vats2023recovering}, have found that pre-training LLMs on datasets processed with privacy-preserving masking does not limit capabilities of models, while privacy benefits are strong. 

\paragraph{Social Learning for LLMs}  \cite{mohtashami2023social} propose the original framework that we expand here. Notable differences from that are \textbf{(i)} our student model can ask for help from the teacher model, \textbf{(ii)} additional teaching algorithms leveraging in-context learning with improved privacy metrics and \textbf{(iii)} showcasing how social learning can enable cascade systems in setups where they would otherwise not be usable.

\paragraph{Synthetic Datasets} LLMs are effective at creating bootstrapping datasets, e.g. by creating task instructions through their own conditional generation~\citep{wang2023selfinstruct}. Similarly, \cite{lee2023rlaif} have shown how alignment data can be synthesized. The student model needs to have such bootstrapping capabilities and the richer this ability is, the better it produces  diverse task transformations that the teacher can better use to explain it back.

\section{Conclusion}
\label{sec:conclusion}

% Quality comparison to baselines
In this paper, we investigated whether LLMs can privately query external LLMs to improve their performance. Indeed, we find that our methods comfortably beat strong baselines that have privacy constraints in place, even with Gemini 1.0 Nano-2 as the student, a 3.5B model that fits on phones.% Fully reaching the teacher's performance is difficult but we show that with Gemini 1.0 Pro, a better student model, we get very close to this.

% Privacy comparison
To evaluate the privacy performance of our methods, we look at two metrics, a simple to interpret count of entities leaked, and another, novel, metric that measures an upper bound of what a curious teacher with auxiliary information could hope to recover from the student's queries. For the first metric, we find masking problems (method 3) to work well, while generating new problems (method 2) with grouping does well in cases where the teacher can be expected to have auxiliary information.

% Conclusion
Ultimately, we note that the choice of methods depends on the concrete threat model considered. For either threat model, we present a compelling system and analysis, which show that leakage can be low while beating strong quality baselines. Additionally, we show how grouping examples improves the privacy metrics, and can, under a given labeling budget constraint, even improve model quality.

Future work in this space could consider more complex forms of student-teacher interactions, further improve the privacy metrics established, and look into modalities other than text.

\section*{Ethics Statement}

% Data minimization enables more private features
Our work supports data minimization principles. It paves the way towards more data staying on users' devices while still offering them intelligent features based on machine learning.

%\section*{Reproducibility Statement}

% Everything should be reproducible
%Our work is based on publicly accessible models and datasets. To make the results easier to reproduce, we release prompts, templates, hyperparameter values and more as part of the supplementary material.

\section*{Acknowledgments}

We would like to express our gratitude towards Matt Sharifi, Tautvydas Misiunas, Hassan Mansoor, Dominik Roblek, Lukas Zilka, Yun Zhu, Jindong (JD) Chen, and Rif A. Saurous for providing crucial feedback on this paper.
All your suggestions greatly improved this paper.
Furthermore, we would also like to thank Amirkeivan Mohtashami whose contributions to the social learning code base remain useful long after his internship.

\bibliography{colm2024_conference}
\bibliographystyle{colm2024_conference}

\section*{Appendix}
\appendix

\section{What Makes for a Good Student and Teacher?}

To decide on promising experiments with our methods, we pose the question: what makes for a good student and teacher combination? We found the following criteria to be useful in deciding on student models and datasets.

\paragraph{Good student} A student model is promising on a dataset, if \textbf{(i)} it can initially not solve the task well (0-shot), but \textbf{(ii)} is able to improve with in-context examples (golden 8-shot). Furthermore, \textbf{(iii)} the student model needs to be able to ask for help via a useful query to the teacher, e.g. it needs to be able to synthesize similar, unlabeled examples.

\paragraph{Good teacher} A teacher model is a good fit for the student model if \textbf{(iv)} it can solve the task much better than the student, meaning even its 0-shot performance is significantly higher than the student's golden 8-shot task success. Furthermore, \textbf{(v)} a good teacher needs to be able to respond to the student's queries, e.g. by providing useful labels for them.

When evaluating whether a new dataset is promising to try with our method, we first check these five criteria.

\section{More Details on the Datasets}
\label{app:datasets}

In this section we provide additional details on the four datasets we use. Table~\ref{fig:template} shows the templates we use for each dataset.

\begin{table*}[h]
    \centering
    %\lstset{linewidth=.6\textwidth,moredelim=**[is][\color{red}\bgroup\texttt{<}\aftergroup>\aftergroup\egroup]{<}{>}}
%\scalebox{.6}{
    \begin{tabular}{c|l}
    \toprule
         Dataset & Example Format \\\midrule
         
         GSM8k &\begin{lstlisting}[basicstyle=\footnotesize]
Question: <question>
Answer: <step-by-step reasoning>
#### <final number>
\end{lstlisting}\\\midrule
         Intent Recognition & \begin{lstlisting}[basicstyle=\footnotesize]
Utterance: <utterance>
Intent: <add_to_playlist, book_restaurant, get_weather,
         play_music, search_screening_event,
         search_creative_work, rate_book>
\end{lstlisting}\\\midrule
Subj & \begin{lstlisting}[basicstyle=\footnotesize]
Text: <text>
Label: <subjective, objective>
\end{lstlisting}\\\midrule
%Object Counting & \begin{lstlisting}[basicstyle=\footnotesize]
%Question: <question>
%Reasoning: <step-by-step reasoning>
%Answer: <answer>
%\end{lstlisting}\\\midrule
Machine Translation & \begin{lstlisting}[basicstyle=\footnotesize]
English sentence: <english sentence>
Basque translation: <basque translation>
\end{lstlisting}\\
\bottomrule
    \end{tabular}
%}
    \caption{The templates for the four datasets we consider. Teacher, student, and baselines can use this information in order to understand how to format examples and where step-by-step reasoning makes sense. This information can either be used in prompts or in constrained decoding configurations.}
    \label{fig:template}
\end{table*}

\paragraph{Grade School Math} GSM8k~\citep{cobbe2021training} is a dataset containing grade school math questions, annotated answers as well as step-by-step reasoning on how to reach the answer. Typical GSM8k examples are written in the form of a story with many entities that we do not want the student to reveal to the teacher.

\paragraph{Intent Recognition} Cascade systems are especially useful for questions that users pose their personal assistant. Intent Recognition~\citep{srivastava2022beyond} is a dataset in which one has to classify an utterance as one of 7 assistant tasks, as shown in Table \ref{fig:template}.

\paragraph{Subj} The Subj dataset~\citep{conneau2018senteval} consists of statements that are either subjective or objective. The model has to classify the statements as one of these two categories.

\paragraph{Machine Translation} LLMs show remarkable machine translation performance. Since performance for high-resource languages is difficult to further improve via ICL, we focus on mid-resource machine translation on the Tatoeba~\citep{tiedemann2020tatoeba} dataset.

\section{Teacher Task Performance}

In Tables~\ref{tab:main-utility} and~\ref{tab:main-privacy} in the main text, we normalize the student's task success by the teacher's performance. In Table~\ref{tab:teacher-utility}, we show this teacher task performance.

\begin{table*}[h]
    \centering
\scalebox{1}{
    \begin{tabular}{c|c|c|c}\toprule
    \textbf{Dataset} & \textbf{Metric} & \textbf{Teacher $n$-shot} & \textbf{Teacher Task Success} \\\midrule
    
\Gape[1ex][1ex]{GSM8k} & accuracy & 0 & 87.8\% \\
\midrule

\makecell{Intent\\Recognition} & accuracy & 0 & 97.4\% \\
\midrule

\Gape[1ex][1ex]{Subj} & accuracy & 8 & 92.3\% \\
\midrule

\makecell{Translation\\en $\rightarrow$ el} & BLEURT & 0 & 90.6\% \\
\bottomrule
    \end{tabular}
}
    \caption{Gemini 1.0 Ultra's task success as the teacher. Even though the teacher itself is not 100\% accurate, the student manages to improve through interaction with the teacher in our experiments. We use 0-shot for the teacher in most experiments, but fall back to 8-shot for Subj since this is a difficult task to do in a 0-shot setting.}
    \label{tab:teacher-utility}
\end{table*}

\section{More Machine Translation Results}
\label{app:mt}

\subsection{Task Performance}
\label{app:mt-performance}

For brevity's sake, we only show results for one language pair in Table~\ref{tab:main-utility} of the main text. Table~\ref{tab:mt-utility} shows the results for all seven languages we consider. Note that each time we translate from English each time since this allows the student model to synthesize useful queries to the teacher even though it does not understand the target language well.

We find our methods to work particularly well for mid-resource languages. Gemini Nano-2 already performs very well on high-resource languages, such as German and Vietnamese, even in the 0-shot setting. Though we do see a small improvement with our methods here, much bigger improvements can be achieved for mid-resource languages.

% MT results: big table with one row per language 
%\newcommand{\smallbold}[1]{\small{\textbf{#1}}}
\begin{table*}[h]
    \centering
\scalebox{0.85}{
    \begin{tabular}{c|c||c|c||c|c|c}\toprule
    \smallbold{From} &
    \smallbold{To} &
    \makecell{\smallbold{Weak Baseline:}\\\smallbold{0-shot}} &
    \makecell{\smallbold{Strong Baseline:}\\\smallbold{Golden Data}\\\smallbold{8-shot}} &
    \makecell{\smallbold{Method 1:}\\\smallbold{Descriptions}\\\smallbold{8-shot}} &
    \makecell{\smallbold{Method 2:}\\\smallbold{New Problems}\\\smallbold{8-shot}} &
    \makecell{\smallbold{Method 3}:\\\smallbold{Replacing}\\\smallbold{8-shot}} \\\midrule
    
\Gape[1ex][1ex]{\small{English}} & \small{German (de)} & 95.4\% &	96.4\% & 92.0\% &	97.6\%	& \textbf{97.3\%} 
\\
\midrule

\Gape[1ex][1ex]{\small{English}} & \small{Greek (el)} & 84.0\% & 88.0\% & 88.3\% & 88.9\% & \textbf{90.7\%}
\\
\midrule

\Gape[1ex][1ex]{\small{English}} & \small{Basque (eu)} & 70.8\% & 72.9\% &	72.8\% & 74.8\% & \textbf{91.0\%}
\\
\midrule

\Gape[1ex][1ex]{\small{English}} & \small{Hebrew (he)} & 81.0\% &	80.9\% &	69.1\% & 80.4\% &	\textbf{86.4\%}
\\
\midrule

\Gape[1ex][1ex]{\small{English}} & \small{Georgian (ka)} & 45.5\% &	46.6\% & 36.3\%	 & 49.7\% &	\textbf{64.4\%}
\\
\midrule

\Gape[1ex][1ex]{\small{English}} & \small{Tagalog (tl)} & 90.8\% &	89.7\% & 	87.6\% 	& 90.9\% & \textbf{94.0\%}
\\
\midrule

\Gape[1ex][1ex]{\small{English}} & \small{Vietnamese (vi)} & 95.7\% &	95.0\% & 90.5\% &	97.5\% &	\textbf{97.1\%} 
\\

\bottomrule
    \end{tabular}
}
    \caption{Machine translation performance (BLEURT) with Gemini 1.0 Nano-2 as the student and Gemini 1.0 Ultra as the teacher. All values are normalized by the teacher's performance. We note that our methods significantly improve results for mid-resource languages while achieving a small improvement for high-resource languages that the student model already understands well.}
    \label{tab:mt-utility}
\end{table*}

\subsection{Qualitative Analysis}
\label{app:mt-qualitative}

To better understand in which cases our techniques improve machine translation, we perform a qualitative analysis, similar to the one in Section~\ref{sec:qualitative}.
Tables~\ref{table:mt-qualitative-eu} and \ref{table:mt-qualitative-el} show the results of these analyses.
We find most error types to significantly decrease with our methods, while the incorrect addition or omission of information slightly increases. 

%Table 3 but once for MT eu and once for MT el.
\begin{table*}[h]
\centering

\def\arraystretch{1.5}
\scalebox{.95}{
\begin{tabular}[h]{c||c||c|c|c} 
\toprule
 \smallbold{Class} &
 \makecell{\smallbold{Strong Baseline:}\\\smallbold{Golden Data}\\\smallbold{8-shot}} &
    \makecell{\smallbold{Method 1:}\\\smallbold{Descriptions}\\\smallbold{8-shot}} &
    \makecell{\smallbold{Method 2:}\\\smallbold{New Problems}\\\smallbold{8-shot}} &
    \makecell{\smallbold{Method 3}:\\\smallbold{Replacing}\\\smallbold{8-shot}}
 \\
\hline \hline
\small Correct translation & 38.4\% &
41.6\% &
40.8\% &
\textbf{64.8\%}
 \\  \hline  \hline
\small Lexical or Semantic error & 50.8\% &
40.4\% &
49.6\% &
\textbf{27.6\%} \\ \hline
\small Grammatical error & 6.0\% &
9.2\% &
\textbf{4.0\%} &
4.8\%
 \\ \hline
\small Contextual or Cultural error &4.0\% &
5.2\% &
3.2\% &
\textbf{0.4\%} 
  \\ \hline
\small Omission or Incorrect Addition & \textbf{0.8\%} &
2.8\% &
2.4\% &
2.4\%
\\ \hline
\small Formatting error & 0.0\% & \ 0.8\% & 0.0\% & 0.0\%
\\
\bottomrule
\end{tabular}
}
\caption{A qualitative error analysis for translation from English to Basque (eu). Lexical, semantic and contextual errors significantly decrease with our methods.}
\label{table:mt-qualitative-eu}
\end{table*}

\begin{table*}[h]
\centering

\def\arraystretch{1.5}
% Don't need scaling right now, but leaving it here since we need to go back to it.
\scalebox{.95}{
\begin{tabular}[h]{c||c||c|c|c} 
\toprule
 \smallbold{Class} &
 \makecell{\smallbold{Strong Baseline:}\\\smallbold{Golden Data}\\\smallbold{8-shot}} &
    \makecell{\smallbold{Method 1:}\\\smallbold{Descriptions}\\\smallbold{8-shot}} &
    \makecell{\smallbold{Method 2:}\\\smallbold{New Problems}\\\smallbold{8-shot}} &
    \makecell{\smallbold{Method 3}:\\\smallbold{Replacing}\\\smallbold{8-shot}}
 \\
\hline \hline
\small Correct translation & 39.0\% &
31.8\% &
41.8\% &
\textbf{50.0\%}
 \\  \hline  \hline
\small Lexical or Semantic error & 39.4\% &
36.1\% &
37.5\% &
\textbf{33.0\%} \\ \hline
\small Grammatical error & 15.3\% &
12.9\% &
14.4\% &
\textbf{10.6\%}
 \\ \hline
\small Contextual or Cultural error & 5.0\% &
12.5\% &
5.5\% &
\textbf{5.0\%} 
  \\ \hline
\small Omission or Incorrect Addition & 1.3\% &
3.2\% &
\textbf{0.9\%} &
1.2\%
\\ \hline
\small Formatting error & \textbf{0.0\%} & \ 3.4\% & \textbf{0.0\%} & 0.2\%
\\
\bottomrule
\end{tabular}
}
\caption{A qualitative error analysis for translation from English to Greek (el). Lexical, semantic and grammatical errors significantly decrease with our methods.}
\label{table:mt-qualitative-el}
\end{table*}

\section{A Student That Is Copying Instead of Learning In-Context}

To evaluate how important ICL is in our setting, we ran additional experiments in which the student copies the teacher's answer instead of learning from it in-context. For the case of \expansionsize $> 1$, the student copies the teacher's most common answer.

We start by noting that such an approach does not satisfy the privacy constraint on many tasks. If a student were for example to achieve high task task success on machine translation by simply copying the teacher's answer, this would imply that the teacher learned the most important parts of the student's original data.

Based on this observation, we stick to GSM8k, intent recognition and Subj for this analysis. To enable the student to achieve a good quality by copying, we rely on the masking approach introduced in Section~\ref{sec:masking}. However, we additionally instruct the student to replace entities in a way that does not change the result. For the case of GSM8k, this means not replacing any numbers and leaving the relationship between any numbers intact.

\begin{table*}[h]
    \centering
\scalebox{1}{
    \begin{tabular}{c|c|c|c}\toprule
    \textbf{Dataset} & \textbf{Student} & \makecell{\textbf{Method 3: Replacing}\\\textbf{with copying}} & \makecell{\textbf{Method 3: Replacing}\\\textbf{with ICL}}  \\\midrule
    
\multirow{2}{*}{GSM8k} & Nano-2 & 9.4\% & \textbf{55.9\%} \\
& Pro & 18.3\% & \textbf{98.3\%}  \\

\midrule

\makecell{Intent\\Recognition} & Nano-2 & 92.7\% & \textbf{94.6\%}  \\
\midrule

\Gape[1ex][1ex]{Subj} & Nano-2 & 74.3\% & \textbf{79.7\%} \\

\bottomrule
    \end{tabular}
}
    \caption{The student learning in-context always outperforms it simply copying the most common label from the teacher. Both methods use 8-shot.}
    \label{tab:copying}
\end{table*}

We find that ICL outperforms copying in our experiments, as shown in Table~\ref{tab:copying}. For intent recognition and Subj, copying works fairly well since there are only a few classes to cover. While most of the time, the examples generated by the student all belong to the same class, there are cases where the original example is close to two similar classes. We find ICL to help in these cases.

For GSM8k copying works much worse. This is even the case when using Pro as a significantly larger student. Looking at experiment logs, the student in this setup struggles to generate queries that do not affect the result.

Based on these results, we decide to stick to ICL for all other experiments, but use the results to influence our Subj prompt.

\section{Example Queries Our Methods Generate}
\label{app:examples}

\newcommand{\MA}[1]{\makecell[l]{#1\\Method 1: Description}}
\newcommand{\MB}[1]{\makecell[l]{#1\\Method 2: New Problem}}
\newcommand{\MC}[1]{\makecell[l]{#1\\Method 3: Masking}}

\begin{table}
  \centering
  \def\arraystretch{1.5}
  \begin{tabularx}{\textwidth}{l|X|X}
    \toprule
    \textbf{Dataset \& Method} & \textbf{Original Example} & \textbf{Student Query} \\
    \midrule
\MA{GSM8k} & Two thirds of Jana's puppies are Pomeranians. One third of the Pomeranians are girls. If there are 6 Pomeranian girls, how many puppies does Jana have? & Division and ratios problem involving percentages. \\ \midrule
\MB{GSM8k} & Two thirds of Jana's puppies are Pomeranians. One third of the Pomeranians are girls. If there are 6 Pomeranian girls, how many puppies does Jana have & If 3/4 of a bag of marbles are red and the rest are blue, and there are 21 red marbles, how many marbles are blue? \\ \midrule
\MC{GSM8k} & Raul had \$87 to spare so he decided to go to the bookshop. Raul bought 8 comics, each of which cost \$4. How much money does Raul have left? & Emily had \$92 to spend at the ice cream shop. She bought 4 ice cream cones, each of which cost \$3. How much money does Emily have left? \\ \midrule
\MC{GSM8k} & The Smith family is getting ready for summer and needs to have their swimming pool filled. The pool company instructed them to measure to find the volume of the pool, then to multiply it by 5.9 to calculate how many gallons of water they need. The cost for the pool company to come and fill the pool is \$0.10 per gallon. Mr. Smith measured and found the pool is 14 feet wide, 25 feet long, and 4 feet deep. How much will it cost to fill the pool? & The Johnson family is getting ready for summer and needs to have their hot tub filled. The hot tub company instructed them to measure to find the volume of the hot tub, then to multiply it by 6.4 to calculate how many gallons of water they need. The cost for the hot tub company to come and fill the hot tub is \$0.15 per gallon. Mr. Johnson measured and found the hot tub is 8 feet wide, 12 feet long, and 3 feet deep. How much will it cost to fill the hot tub \\ \midrule
\MC{Intent Recognition} & Show the TV show “The Last Samurai." & Play the song 'Blinding Lights.'  \\ \midrule
\MC{Subj} & 'seven years later , alan is given a second chance , a reunion with tommy and the opportunity to set things right . & five years later , jesse is given a second chance , a reunion with thomas and the opportunity to set things right . \\ \midrule
\MB{Translation} & I bought this printer yesterday & I went to the store yesterday to buy some groceries \\ \midrule
\MC{Translation} & Tom bought three dozen eggs & Jane bought five dozen apples \\
    \bottomrule
    \end{tabularx}
  \caption{Examples where the student generates good queries.}
  \label{tab:examples-good}
\end{table}%

\begin{table}
  \centering
  \def\arraystretch{1.5}
  \begin{tabularx}{\textwidth}{l|X|X}
    \toprule
    \textbf{Dataset \& Method} & \textbf{Original Example} & \textbf{Student Query} \\
    \midrule
\MA{GSM8k} & Sandra, the florist around the corner, is very unhappy with Fred's incomplete order delivery. Sandra had ordered four times as many red roses as white carnations. She also ordered 200 pink calla lilies, which were five times the number of white carnations. Sandra has threatened to switch suppliers if the missing red roses are not delivered by 5 pm. To keep Sandra's business, how many red roses must Fred deliver by 5 pm &
This is a problem of determining the number of red roses Fred must deliver in order to appease Sandra \\ \midrule
\MB{GSM8k} & James has to buy insurance.  Since he had an accident it was 60\% more than normal.  The normal cost is \$120 a month.  How much does he pay a year? & John had an accident that was 75\% more expensive than usual.  If the usual monthly cost was \$150, how much did he pay for insurance for the year? \\ \midrule
\MC{GSM8k} & Raul had \$87 to spare so he decided to go to the bookshop. Raul bought 8 comics, each of which cost \$4. How much money does Raul have left? & Olivia had \$222 to spare so she decided to go to the pet shop. Olivia bought 28 dogs, each of which cost \$15. How much money does Olivia have left? \\ \midrule
\MC{Intent Recognition} & Rate “Awaiting Strange Gods: Weird and Lovecraftian Fictions” a 1. & Rate "Awaiting Strange Gods: Weird and Lovecraftian Fictions" a 10. It was so bad I wanted to cry and rip my eyes out. Even the sound of the ending credits was painful to endure. \\ \midrule
\MB{Subj} & 'adrien , a former high-flying bank robber , recently released from prison and suffering from an incurable form of cancer . . .' & adrien , a former high-flying bank robber , was recently released from prison after serving a long sentence for his crimes . . \\ \midrule
\MA{Translation} & I went aboard & Translate the English phrase "I went aboard" into Greek.
\\ \bottomrule
    \end{tabularx}
  \caption{Examples in which the student leaks information or does not generate a useful query.}
  \label{tab:examples-bad}
\end{table}%

Table~\ref{tab:examples-good} shows example student problems and queries that work well. In all of these examples, the student is able to generate a query to the teacher that does not verbatim leak sensitive information but that nevertheless allows the teacher to respond with useful examples.

In Table ~\ref{tab:examples-bad}, we show examples in which the student does not generate a good query. In most of these cases, the student leaks sensitive information. In some, the student generates a query that does not make sense.

\end{document}